\documentclass{article}
\usepackage[utf8]{inputenc}
\usepackage{hyperref}
\usepackage[T1]{fontenc}
\usepackage{geometry}
\usepackage{amsmath, amssymb}
\usepackage{bm}           
\usepackage{graphicx}
\usepackage{booktabs}     
\usepackage[ruled,vlined,linesnumbered]{algorithm2e} 

\title{Spectral-Window Hybrid (SWH)}
\author{
  \textbf{Vladimer Khasia} \\
  Independent Researcher \\
  \texttt{vladimer.khasia.1@gmail.com}
}
\date{January 3, 2026}

\begin{document}

\maketitle

\begin{abstract}
Scaling sequence modeling to extreme contexts requires balancing computational efficiency with representational expressivity. While Transformers provide precise retrieval via the attention mechanism, their quadratic $\mathcal{O}(T^2)$ complexity limits their application to long-horizon tasks. In this work, we propose the \textbf{Spectral-Window Hybrid (SWH)}, an architecture that decouples sequence modeling into two \textit{parallel} streams: a global branch utilizing the Convolution Theorem to model long-range decay dynamics in $\mathcal{O}(T \log T)$ time, and a local branch employing sliding-window attention for token interactions within a bounded context. By aggregating these representations, SWH avoids the computational bottleneck of global attention while retaining local precision. We demonstrate that SWH matches the perplexity of standard Transformers on short contexts while enabling efficient linear scaling to extended sequences. The code is available at {\url{https://github.com/VladimerKhasia/SWH}}
\end{abstract}

\section{Introduction}

Sequence modeling constitutes the backbone of modern Artificial Intelligence, driving progress in Large Language Models (LLMs). The Transformer architecture \cite{Vaswani2017AttentionIA} serves as the standard, primarily due to the effectiveness of the self-attention mechanism in modeling dense, pairwise dependencies. However, standard global self-attention incurs a computational cost of $\mathcal{O}(T^2)$ in both time and memory. While hardware-aware optimizations like FlashAttention \cite{Dao2022FlashAttentionFA} reduce activation memory complexity to linear and improve computational throughput, the asymptotic quadratic \textit{compute} scaling remains a bottleneck for contexts spanning millions of tokens.

To address this scaling challenge, research has explored \textit{Structured State Space Models (SSMs)}. Early Linear Time Invariant (LTI) systems, such as S4 \cite{Gu2021EfficientlyML}, leverage continuous-time systems that admit a convolutional representation to achieve linear scaling. More recently, Selective SSMs like Mamba \cite{Gu2023MambaLS} introduced input-dependent dynamics to address the limitations of LTI systems in content-dependent reasoning and in-context learning.

Building on these advancements, recent architectures have moved toward \textit{Hybrid} models that combine the efficiency of recurrence with the precision of attention. Jamba \cite{Lieber2024JambaAH} utilizes a sequential hybrid approach, interleaving blocks of Mamba layers with occasional global attention layers to balance throughput with long-range retrieval. Similarly, Griffin \cite{De2024GriffinMG} combines recurrent blocks (specifically the RG-LRU) with local sliding-window attention, avoiding global attention entirely to maintain fixed-size inference states.

In this work, we explore an alternative arrangement to these sequential hybrids. We propose the \textbf{Spectral-Window Hybrid (SWH)}, an architecture that utilizes a \textit{parallel decoupled} approach of global convolution and local attention. 

\section{Methodology}
\label{sec:methodology}

We propose the \textbf{Spectral-Window Hybrid (SWH)} architecture, which decomposes sequence modeling into two parallel processes: (1) a global deterministic Linear Time-Invariant (LTI) system solved via the Convolution Theorem, and (2) a local probabilistic attention mechanism with sliding window memory.

Let $\mathbf{X} \in \mathbb{R}^{B \times T \times D}$ denote the input tensor, where $B$ is the batch size, $T$ is the sequence length, and $D$ is the embedding dimension. The output $\mathbf{Y}$ is computed as the projection of the sum of the spectral and local branches\footnote{Bias terms are omitted in the mathematical formulation for brevity. In our implementation, bias terms are explicitly disabled for stability in all linear projections, with the exception of the spectral branch input projection.}:

\begin{equation}
    \mathbf{Y} = (\text{RMSNorm}(\mathbf{Y}_{\text{spec}}) + \mathbf{Y}_{\text{local}})\mathbf{W}_{\text{out}}
\end{equation}

where $\mathbf{W}_{\text{out}} \in \mathbb{R}^{D \times D}$ is a learnable projection and $\mathbf{Y}_{\text{spec}}, \mathbf{Y}_{\text{local}}$ are the outputs of the respective branches defined below.

\subsection{Global Branch: Causal Spectral Convolution}

We model global dependencies using a parameterized causal convolution. We define a learnable kernel $\mathbf{K} \in \mathbb{R}^{T \times D}$ parameterized as the impulse response of a damped harmonic oscillator. For a time step $t \in [0, T-1]$ and channel $c \in [0, D-1]$:

\begin{equation}
    K_{t,c} = \exp(-\lvert \alpha_c \rvert \cdot t) \cos(\omega_c \cdot t)
\end{equation}

where $\bm{\alpha}, \bm{\omega} \in \mathbb{R}^D$ are learnable decay and frequency parameters. The spectral output is obtained via depth-wise causal convolution:

\begin{equation}
    \mathbf{Y}_{\text{spec}} = (\mathbf{X}\mathbf{W}_{\text{conv}}) * \mathbf{K}
\end{equation}

where $\mathbf{W}_{\text{conv}} \in \mathbb{R}^{D \times D}$ is the input projection. 

\subsubsection{Efficient Implementation via FFT}
Direct computation of Eq.~(3) requires $\mathcal{O}(T^2)$ operations. We utilize the Convolution Theorem for $\mathcal{O}(T \log T)$ complexity. To enforce causality and prevent circular aliasing inherent to the Discrete Fourier Transform (DFT), the signal and kernel must be zero-padded along the time dimension to length $L \geq 2T$. Computations in the spectral domain are performed in single-precision (FP32) to ensure numerical stability.

Let $\mathcal{F}$ and $\mathcal{F}^{-1}$ denote the FFT and inverse FFT. We compute:

\begin{equation}
    \mathbf{Y}_{\text{spec}} = \text{Crop}_{0:T-1}\left( \mathcal{F}^{-1}\big( \mathcal{F}(\text{Pad}(\mathbf{X}\mathbf{W}_{\text{conv}}, 2T)) \odot \mathcal{F}(\text{Pad}(\mathbf{K}, 2T)) \big) \right)
\end{equation}

where $\odot$ denotes the Hadamard product broadcast over the batch dimension, and cropping extracts the first $T$ time steps.

\subsection{Local Branch: Chunked Sliding-Window Attention}

Local token interactions are recovered via a sliding window attention mechanism of size $W$. To optimize memory access patterns, we formulate this as \textbf{Chunked Attention}.

The input sequence is partitioned into $N = \lceil T/W \rceil$ non-overlapping chunks. If $T$ is not divisible by $W$, the sequence is right-padded with zeros to ensure strict partitioning. Let $\mathbf{X}^{(n)} \in \mathbb{R}^{B \times W \times D}$ denote the $n$-th chunk. We compute queries $\mathbf{Q}^{(n)}$, keys $\mathbf{K}^{(n)}$, and values $\mathbf{V}^{(n)}$ via linear projections.

\subsubsection{Rotary Position Embeddings}
To preserve relative positional information, we apply Rotary Positional Embeddings (RoPE) to the \textbf{global} query and key sequences prior to partitioning. Let $f_R(\cdot, m)$ be the rotary injection function for absolute position $m$. For a token at global index $t$, the transformed states are:

\begin{equation}
    \tilde{\mathbf{q}}_{t} = f_R(\mathbf{q}_{t}, t), \quad \tilde{\mathbf{k}}_{t} = f_R(\mathbf{k}_{t}, t)
\end{equation}

These global tensors are subsequently split into chunks $\tilde{\mathbf{Q}}^{(n)}, \tilde{\mathbf{K}}^{(n)}$ for attention computation.

\subsubsection{Context Management and Masking}
For the $n$-th chunk, the key-value context consists of the previous chunk (cached) and the current chunk. This defines a sliding window of size $2W$ (effective receptive field):

\begin{equation}
    \mathbf{K}_{\text{ctx}}^{(n)} = [\mathbf{K}^{(n-1)}; \mathbf{K}^{(n)}], \quad \mathbf{V}_{\text{ctx}}^{(n)} = [\mathbf{V}^{(n-1)}; \mathbf{V}^{(n)}]
\end{equation}

where $[\cdot ; \cdot]$ denotes concatenation along the time dimension. For $n=1$, $\mathbf{K}^{(0)}, \mathbf{V}^{(0)}$ are initialized as zero tensors and logically treated as padding.

The attention logits $\mathbf{S}^{(n)} \in \mathbb{R}^{B \times H \times W \times 2W}$ are computed as:

\begin{equation}
    \mathbf{S}^{(n)} = \frac{\tilde{\mathbf{Q}}^{(n)} (\tilde{\mathbf{K}}_{\text{ctx}}^{(n)})^\top}{\sqrt{D/H}} + \mathbf{M}^{(n)}
\end{equation}

where $H$ is the number of heads and $\mathbf{M}^{(n)}$ is the \textbf{Block-Causal Mask}. For a generic chunk $n > 1$, the previous chunk ($j \in [1, W]$) is fully visible, while the current chunk ($j \in [W+1, 2W]$) requires causal masking. For the first chunk ($n=1$), the previous context is invalid and must be masked out to prevent probability leakage. The mask is defined as:

\begin{equation}
    M_{i,j}^{(n)} = \begin{cases} 
    -\infty & \text{if } n=1 \text{ and } j \leq W \quad \text{(Invalid Context)} \\
    0 & \text{if } n > 1 \text{ and } j \leq W \quad \text{(Valid Context)} \\
    0 & \text{if } W < j \leq W+i \quad \text{(Current Chunk, Causal)} \\
    -\infty & \text{otherwise}
    \end{cases}
\end{equation}

The final local output is obtained by projecting the concatenated heads:

\begin{equation}
    \mathbf{Y}_{\text{local}} = \text{Concat}_{n=1}^N \left( \text{Softmax}(\mathbf{S}^{(n)}) \mathbf{V}_{\text{ctx}}^{(n)} \right) \mathbf{W}_{\text{local}}
\end{equation}

where $\mathbf{W}_{\text{local}} \in \mathbb{R}^{D \times D}$ is the output projection matrix.

\subsection{Algorithm}

The forward pass implementation is detailed in Algorithm \ref{alg:swh}.

\begin{algorithm}[h!]
\caption{Spectral-Window Hybrid (SWH) Forward Pass}
\label{alg:swh}
\SetAlgoLined
\SetKwInOut{Input}{Input}
\SetKwInOut{Output}{Output}
\SetKwFunction{FFT}{rFFT}
\SetKwFunction{IFFT}{irFFT}
\SetKwFunction{RMS}{RMSNorm}
\SetKwFunction{Softmax}{Softmax}
\SetKwFunction{Concat}{Concat}
\SetKwFunction{Pad}{Pad}
\SetKwFunction{Crop}{Crop}
\SetKwFunction{RoPE}{ApplyRoPE}

\Input{$\mathbf{X} \in \mathbb{R}^{B \times T \times D}$, Decay $\bm{\alpha}$, Freq $\bm{\omega}$, Window $W$}
\Output{$\mathbf{Y} \in \mathbb{R}^{B \times T \times D}$}

\BlankLine
\tcc{\textbf{Branch 1: Causal Spectral Convolution}}
$\mathbf{U} \leftarrow \mathbf{X}\mathbf{W}_{\text{conv}}$\;
$K_{t,c} \leftarrow e^{-|\alpha_c| t} \cos(\omega_c t)$ for $t \in [0, T-1]$\;
\tcc{Zero-pad to $2T$ for linear convolution via FFT (FP32)}
$\hat{\mathbf{U}} \leftarrow \Pad(\mathbf{U}, 2T)$; $\hat{\mathbf{K}} \leftarrow \Pad(\mathbf{K}, 2T)$\;
$\mathbf{H}_{\text{spec}} \leftarrow \IFFT\big( \FFT(\hat{\mathbf{U}}) \odot \FFT(\hat{\mathbf{K}}) \big)$ \tcc*[r]{$\odot$ is broadcast over B}
$\mathbf{Y}_{\text{spec}} \leftarrow \mathbf{H}_{\text{spec}}[0:T-1]$ \tcc*[r]{Crop to length T}

\BlankLine
\tcc{\textbf{Branch 2: Chunked Sliding-Window Attention}}
$\mathbf{Q}, \mathbf{K}, \mathbf{V} \leftarrow \mathbf{X}\mathbf{W}_Q, \mathbf{X}\mathbf{W}_K, \mathbf{X}\mathbf{W}_V$\;
$\mathbf{Q}, \mathbf{K} \leftarrow \RoPE(\mathbf{Q}, \mathbf{K})$ \tcc*[r]{Global Position Injection}

\tcc{Handle padding if $T$ is not divisible by $W$}
$P \leftarrow (W - (T \pmod W)) \pmod W$\;
$\mathbf{Q}, \mathbf{K}, \mathbf{V} \leftarrow \Pad(\mathbf{Q}, \mathbf{K}, \mathbf{V}, P)$\;

Split $\mathbf{Q}, \mathbf{K}, \mathbf{V}$ into $N$ chunks of size $W$\;
Initialize buffers $\mathbf{K}_{\text{prev}}, \mathbf{V}_{\text{prev}} \leftarrow \mathbf{0}^{B \times W \times D}$\;
$\mathbf{Y}_{\text{att}} \leftarrow []$\;

\For{$n \leftarrow 1$ \KwTo $N$}{
    $\mathbf{K}_{\text{ctx}} \leftarrow \Concat(\mathbf{K}_{\text{prev}}, \mathbf{K}^{(n)})$; $\mathbf{V}_{\text{ctx}} \leftarrow \Concat(\mathbf{V}_{\text{prev}}, \mathbf{V}^{(n)})$\;
    
    Calculate Scores $\mathbf{A}^{(n)}$ via Eq. (7)\;
    Apply Block-Causal Mask $\mathbf{M}^{(n)}$ defined in Eq. (8)\;
    \If{$n=1$}{
        Mask region $[0:W]$ with $-\infty$ \tcc*[r]{Ignore padding}
    }
    
    $\mathbf{O}^{(n)} \leftarrow \Softmax(\mathbf{A}^{(n)}) \mathbf{V}_{\text{ctx}}$\;
    Append $\mathbf{O}^{(n)}$ to $\mathbf{Y}_{\text{att}}$\;
    
    $\mathbf{K}_{\text{prev}} \leftarrow \mathbf{K}^{(n)}$; $\mathbf{V}_{\text{prev}} \leftarrow \mathbf{V}^{(n)}$\;
}
$\mathbf{Y}_{\text{local}} \leftarrow \Concat(\mathbf{Y}_{\text{att}})$\;
\If{$P > 0$}{
    $\mathbf{Y}_{\text{local}} \leftarrow \Crop(\mathbf{Y}_{\text{local}}, 0, T-1)$ \tcc*[r]{Remove padding}
}
$\mathbf{Y}_{\text{local}} \leftarrow \mathbf{Y}_{\text{local}}\mathbf{W}_{\text{local}}$ \tcc*[r]{Output Projection}

\BlankLine
\tcc{\textbf{Fusion}}
$\mathbf{Y} \leftarrow (\RMS(\mathbf{Y}_{\text{spec}}) + \mathbf{Y}_{\text{local}})\mathbf{W}_{\text{out}}$\;
\Return{$\mathbf{Y}$}
\end{algorithm}

\subsection{Complexity Analysis}

We compare the asymptotic complexity of SWH against the standard Transformer (Baseline). Let $T$ be the sequence length and $D$ be the hidden dimension.

\begin{table}[h!]
\centering
\begin{tabular}{lcc}
\toprule
\textbf{Method} & \textbf{Time Complexity} & \textbf{Space Complexity} \\
\midrule
Standard Attention & $\mathcal{O}(T^2 D)$ & $\mathcal{O}(T^2 + TD)$ \\
Spectral Branch & $\mathcal{O}(T \log T \cdot D)$ & $\mathcal{O}(T D)$ \\
Window Attention & $\mathcal{O}(T \cdot W \cdot D)$ & $\mathcal{O}(T \cdot W \cdot H)$ \\
\textbf{SWH (Total)} & $\mathbf{\mathcal{O}(T( \log T + W) D)}$ & $\mathbf{\mathcal{O}(TD)}$ \\
\bottomrule
\end{tabular}
\caption{Complexity comparison. $W$ is fixed window size ($W \ll T$).}
\label{tab:complexity}
\end{table}

As shown in Table \ref{tab:complexity}, the SWH architecture eliminates the quadratic bottleneck $\mathcal{O}(T^2)$. The spectral branch scales log-linearly due to the FFT, while the window attention scales linearly with $T$ given a constant $W$.

\section{Experiments}
\label{sec:experiments}

We evaluate the proposed Spectral-Window Hybrid (SWH) architecture against a standard Transformer baseline. Our evaluation proceeds in two phases: (1) a preliminary synthetic evaluation to verify fundamental sequence modeling capabilities and extrapolation potential, and (2) a large-scale language modeling objective to assess performance on real-world data. Finally, we analyze the computational efficiency regarding latency and memory usage.

All experiments were conducted on Kaggle using two NVIDIA T4 GPUs. 

\subsection{Preliminary Synthetic Evaluation}
\label{subsec:synthetic}

Before scaling to natural language, we assessed the method's ability to capture essential algorithmic primitives. We compared a basic SWH method against a standard causal Attention (both with $D=128$, $L=2$, $H=4$) on five diagnostic tasks. The models were trained for 3,000 steps. The tasks included:
\begin{itemize}
    \item \textbf{Associative Recall:} Retrieving a value associated with a specific key.
    \item \textbf{Induction Head:} A core mechanism for in-context learning.
    \item \textbf{Sorting:} Outputting the sorted version of an integer sequence.
    \item \textbf{Length Generalization (LenGen):} Trained on sequence length $T=32$, evaluated on $T=128$.
    \item \textbf{Needle-in-a-Haystack:} Retrieving a specific token pair buried in a long sequence (evaluated at $T=256$ after training on $T=32$).
\end{itemize}

\begin{figure}[h!]
    \centering
    \includegraphics[width=0.9\linewidth]{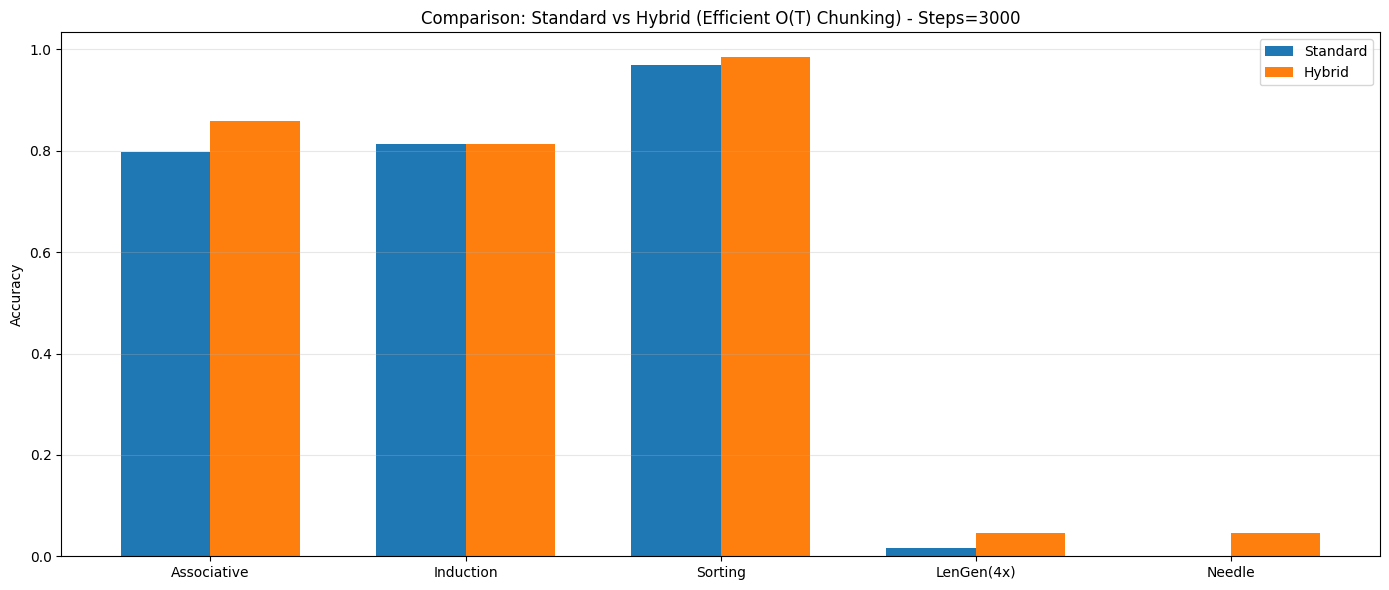}
    \caption{\textbf{Synthetic Task Performance.} Comparison of Standard Attention vs. SWH on five algorithmic primitives. SWH demonstrates superior generalization on out-of-distribution lengths (LenGen, Needle) while maintaining strong performance on in-distribution tasks.}
    \label{fig:synthetic_bar}
\end{figure}

The results (Table \ref{tab:synthetic_results} and Figure \ref{fig:synthetic_bar}) indicate that SWH matches the baseline on in-distribution tasks (Associative, Induction, Sorting). Notably, SWH exhibits superior capability in length extrapolation. While the baseline fails completely on the Needle task ($0.00$ accuracy) when the sequence length increases beyond training limits, SWH maintains non-zero performance ($0.05$), attributed to the continuous nature of the spectral decay and the localized sliding window.

\begin{table}[h!]
\centering
\caption{Accuracy on Synthetic Primitives (Training Steps=3000). SWH matches the baseline on standard tasks and shows superior capabilities in length generalization.}
\label{tab:synthetic_results}
\begin{tabular}{lccccc}
\toprule
\textbf{Method} & \textbf{Associative} & \textbf{Induction} & \textbf{Sorting} & \textbf{LenGen (4x)} & \textbf{Needle} \\
\midrule
Standard Transformer & 0.80 & 0.81 & 0.97 & 0.02 & 0.00 \\
\textbf{SWH (Ours)} & \textbf{0.86} & \textbf{0.81} & \textbf{0.98} & \textbf{0.05} & \textbf{0.05} \\
\bottomrule
\end{tabular}
\end{table}

\subsection{Language Modeling on FineWeb-Edu}
\label{subsec:lm_experiment}

We trained a 125M parameter class model on the \texttt{FineWeb-Edu} dataset (HuggingFaceFW/fineweb-edu sample-10BT \cite{lozhkov2024fineweb-edu}). Both the Baseline and SWH configurations utilized $L=12$ layers, $D=768$, and $H=12$ heads. Training was conducted with a block size of 1024, a batch size of 64 (accumulated), and a learning rate of $6 \times 10^{-4}$ with cosine decay.

\begin{figure}[h!]
    \centering
    \includegraphics[width=0.8\linewidth]{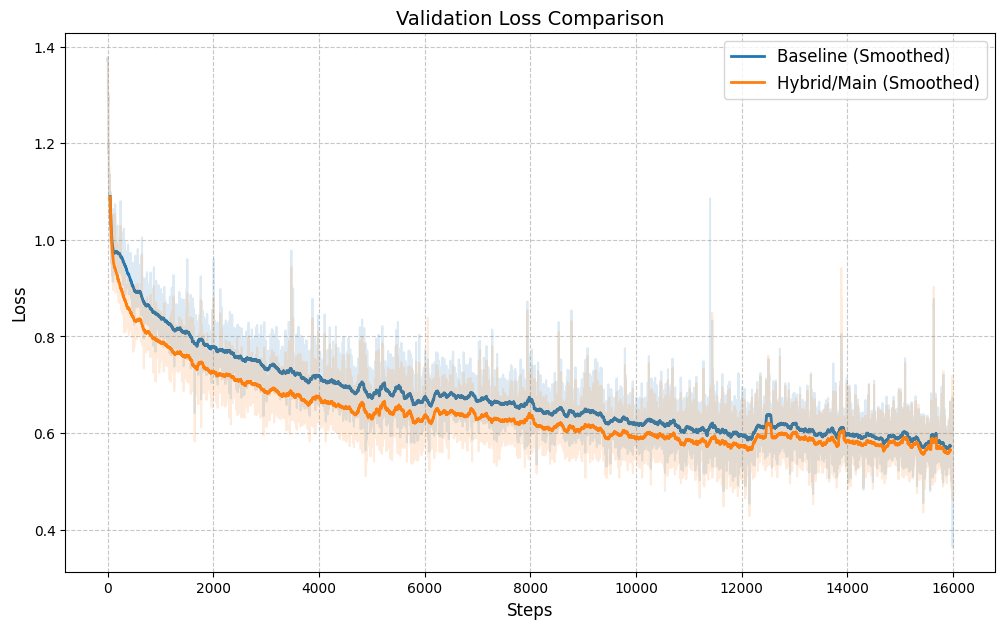}
    \caption{\textbf{Validation Loss on FineWeb-Edu.} The SWH model (Orange) consistently achieves lower loss compared to the Standard Transformer Baseline (Blue). Smoothed curves (solid lines) are overlaid on raw loss values (transparent).}
    \label{fig:loss_curve}
\end{figure}

Figure \ref{fig:loss_curve} illustrates the validation loss over 16,000 steps. The SWH architecture achieves a consistently lower perplexity compared to the Transformer baseline. The hybrid nature of the architecture allows it to capture both long-range global contexts (via the spectral branch) and sharp local dependencies (via the window attention), resulting in improved convergence efficiency.

\subsection{Computational Efficiency}
\label{subsec:efficiency}

A primary motivation for SWH is breaking the quadratic bottleneck of standard attention. We benchmarked the inference latency and VRAM usage across sequence lengths $T \in [512, 1024, 2048, 4096]$.

\begin{figure}[h!]
    \centering
    \includegraphics[width=1.0\linewidth]{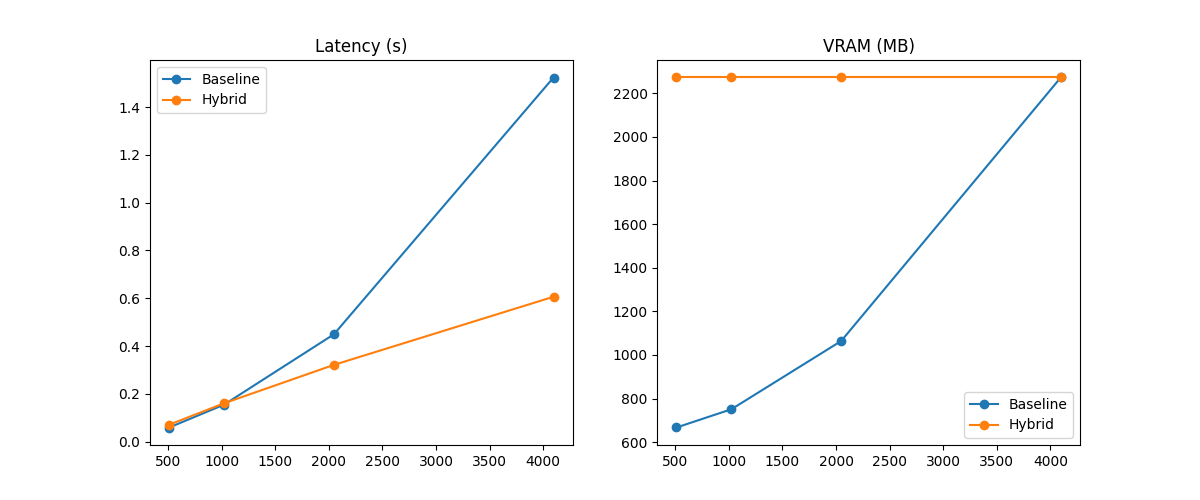}
    \caption{\textbf{Efficiency Analysis.} Left: Inference latency (seconds) vs. sequence length. Right: VRAM usage (MB) vs. sequence length. SWH displays linear scaling, whereas the baseline exhibits quadratic growth, making SWH significantly more efficient for long sequences.}
    \label{fig:efficiency}
\end{figure}

As shown in Figure \ref{fig:efficiency}, the Standard Transformer exhibits quadratic scaling $\mathcal{O}(T^2)$, resulting in a latency of $>1.4$s and memory saturation at $T=4096$. In contrast, SWH scales linearly. At $T=4096$, SWH reduces latency by approximately $60\%$ ($0.6$s) and maintains constant memory usage due to the fixed-size window attention and the $\mathcal{O}(T)$ memory cost of the spectral convolution.

\section{Conclusion}
\label{sec:conclusion}

In this paper, we introduced the Spectral-Window Hybrid (SWH), a sequence modeling architecture designed to reconcile the efficiency of global convolutions with the precision of local attention. By decomposing the modeling task into parallel spectral and windowed-spatial components, SWH eliminates the $\mathcal{O}(T^2)$ complexity of standard Transformers, achieving near-linear scaling via FFT-based convolutions. 

This architecture offers a scalable pathway for long-context modeling, providing the inference speed of linear models with the local representational power of Transformers.

\clearpage
\appendix

\section{Experimental Details}

\subsection{Hyperparameters}
Table \ref{tab:hyperparams} details the hyperparameters used for the main Language Modeling experiment described in Section \ref{subsec:lm_experiment}.

\begin{table}[h!]
\centering
\caption{Hyperparameters for FineWeb-Edu Training.}
\label{tab:hyperparams}
\begin{tabular}{lc}
\toprule
\textbf{Parameter} & \textbf{Value} \\
\midrule
Parameter Count & $\approx 125$M \\
Layers ($L$) & 12 \\
Hidden Dimension ($D$) & 768 \\
Heads ($H$) & 12 \\
Block Size ($T$) & 1024 \\
Vocab Size & 50304 \\
\midrule
Optimizer & AdamW \\
Learning Rate & $6 \times 10^{-4}$ \\
Weight Decay & $0.1$ \\
Batch Size (Global) & 64 \\
Gradient Accumulation & 8 steps \\
Precision & Mixed (BF16/FP16) \\
FFT Precision & FP32 \\
\bottomrule
\end{tabular}
\end{table}

\subsection{Synthetic Task Definitions}
The synthetic tasks used in Section \ref{subsec:synthetic} are defined as follows:

\begin{itemize}
    \item \textbf{Associative Recall:} Given a sequence of key-value pairs (e.g., $k_1, v_1, k_2, v_2, \dots$), the model must predict $v_i$ when prompted with $k_i$ at the end of the sequence.
    \item \textbf{Induction Head:} The model must complete a pattern $A \dots B \dots A \to B$. A token $A$ appears, followed eventually by $B$. When $A$ appears again later, the model must predict $B$.
    \item \textbf{Sorting:} The input is a sequence of random integers drawn from the vocabulary. The target is the same sequence sorted in ascending order.
    \item \textbf{Length Generalization (LenGen):} The model is trained on standard associative recall with sequence length $T_{\text{train}}$. It is evaluated on sequences of length $T_{\text{test}} = 4 \times T_{\text{train}}$.
    \item \textbf{Needle-in-a-Haystack:} A specific key-value pair ("the needle") is inserted at a random position within a long sequence ("the haystack") of noise tokens. The model is queried for the value at the very end.
\end{itemize}

\bibliographystyle{plain}
\bibliography{references} 
\end{document}